\documentclass[times, review, 10pt]{elsarticle}

\usepackage{amssymb}
\usepackage{amsmath}

\usepackage{graphicx}
\usepackage{caption}
\usepackage{wrapfig}
\usepackage{tabularx}
\usepackage{multicol}
\usepackage{array}
\usepackage{xspace}
\usepackage{makecell}
\usepackage{tabu}
\usepackage{xfrac}
\usepackage{color}
\usepackage{float}
\usepackage{colortbl}
\usepackage{nicefrac}
\usepackage{calc}
\usepackage{nccmath}
\usepackage{amsmath}
\usepackage{mathtools}
\usepackage{amsthm}
\usepackage{booktabs}
\usepackage[table]{xcolor}
\usepackage{hyperref}

\usepackage{algorithm}
\usepackage{algorithmicx}
\usepackage{algpseudocode}
\usepackage{setspace} 
\usepackage{caption}

\usepackage{booktabs}
\usepackage{multirow}

\usepackage[capitalize]{cleveref}
\crefname{section}{Sec.}{Secs.}
\Crefname{section}{Section}{Sections}
\Crefname{table}{Table}{Tables}
\crefname{table}{Tab.}{Tabs.}

\newcommand{\figref}[1]{Fig.~\ref{#1}}
\newcommand{\tabref}[1]{Table~\ref{#1}}

\definecolor{CuGray}{gray}{0.9}
\newcolumntype{g}{>{\columncolor{CuGray}}c}

\makeatletter
\DeclareRobustCommand\onedot{\futurelet\@let@token\@onedot}
\def\@onedot{\ifx\@let@token.\else.\null\fi\xspace}

\newcommand{\dashrule}[1][black]{%
  \color{#1}\rule[\dimexpr.5ex-.2pt]{4pt}{.4pt}\xleaders\hbox{\rule{4pt}{0pt}\rule[\dimexpr.5ex-.2pt]{4pt}{.4pt}}\hfill\kern0pt%
}

\def\eg{\emph{e.g}\onedot}

\def\etal{\emph{et~al}\onedot}

\journal{Pattern Recognition}

\begin{document}

\begin{frontmatter}



\title{CLIP Can Understand Depth}


\author[1]{Sohee Kim\fnref{equal1}}
\author[1]{Jisu Kang\fnref{equal1}}
\author[1]{Dunam Kim\fnref{equal1}}
\author[1]{Seokju Lee\corref{cor1}}
\ead{slee@kentech.ac.kr}
\cortext[cor1]{Corresponding author.}
\fntext[equal1]{These authors contributed equally to this work.}

\affiliation[1]{
                organization={Korea Institute of Energy Technology (KENTECH)},
                country={Republic of Korea}
                }

\begin{abstract}
In this paper, we demonstrate that CLIP can also be adapted to downstream tasks where its vision-language alignment is suboptimally learned during pre-training on web-crawled data, all without requiring fine-tuning. We explore the case of monocular depth estimation, where CLIP's contrastive prior struggles to generalize, compared to its success in domains such as generative modeling and semantic segmentation. Since CLIP fails to consistently capture similarities between image patches and natural language prompts describing distance, we eliminate the use of its pre-trained natural language token embeddings and distill the semantic prior of its frozen text encoder into a single learnable embedding matrix called \emph{``mirror''}. The main design goal of \emph{mirror} is to derive a non-human language prompt that approximates an optimal natural language prompt: ``\emph{How far is this location from the camera?}'' Using this approach, we jointly train two lightweight modules, a \emph{mirror} and a compact decoder, on top of a frozen CLIP for dense depth prediction. Compared to conventional depth models, our framework is significantly more efficient in terms of parameters and computation. The resulting model exhibits impressive performance, matching several state-of-the-art vision models on the NYU Depth v2 and KITTI benchmark datasets, while outperforming all vision-language depth models based on a frozen CLIP prior. Specifically, our method reduces the Absolute Relative Error (Abs Rel) by 68.7\% on NYU Depth v2 and by 75.6\% on KITTI compared to the method of Auty \etal, a representative CLIP-based baseline. Experiments demonstrate that the suboptimal depth understanding of CLIP in terms of spatial and temporal consistency can be significantly corrected without either fine-tuning it or concatenating \emph{mirror} with its pre-trained subword token embeddings. Furthermore, an ablation study on the convergence status of \emph{mirror} shows that it is implicitly trained to capture objects, such as humans and windows, where semantic cues play an important role in detection.

\end{abstract}




\end{frontmatter}

\section{Introduction}

\begin{figure*}[!t]
\centering
\includegraphics[width=1.0\linewidth]{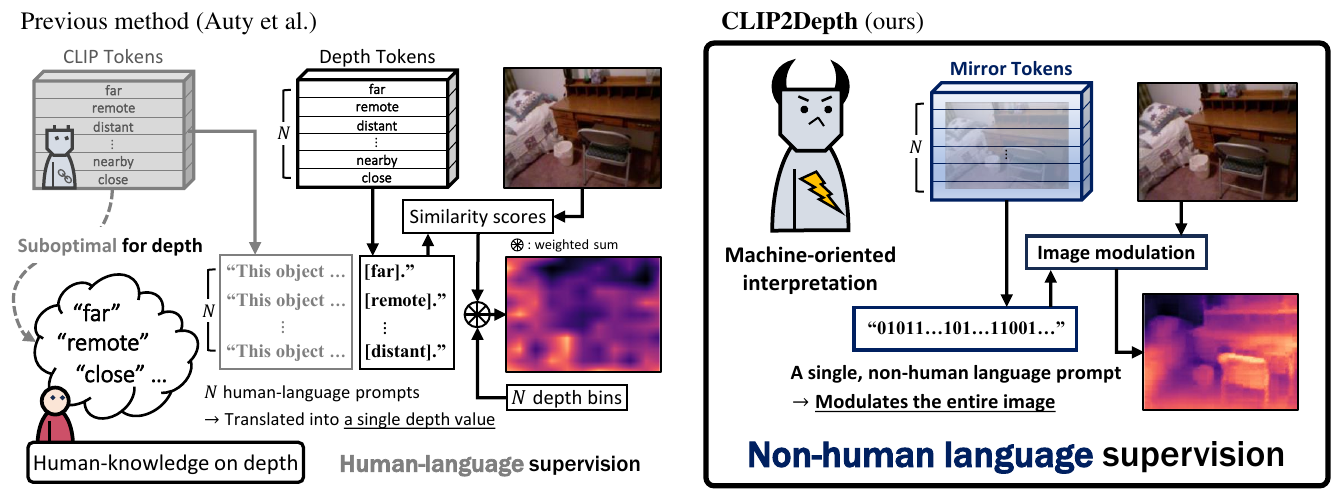}    
\caption{
  \textbf{The CLIP2Depth framework:} Training monocular depth with non-human language supervision utilizing the \emph{mirror.}
  For clarity, our goal is not to propose a novel depth-specific method but rather to demonstrate that CLIP can be adapted to tasks where its vision-language alignment is suboptimally pretrained, all without fine-tuning.}
\label{fig:teaser}
\end{figure*}

Radford~\etal~\cite{radford2021learning} demonstrated the remarkable generalization ability of CLIP, opening a novel research direction in deep learning based on vision-language models. However, despite its broad applicability, CLIP shows clear limitations in domains that were suboptimally represented during its pretraining. 
A natural solution is to fine-tune CLIP for specific downstream tasks while preserving its contrastive prior; however, this often incurs significant computational and data costs. 
Unlike conventional monocular depth estimation methods that rely solely on visual features, fine-tuning CLIP offers a distinct path to generalization by adapting vision-language representations to depth prediction. 
This paradigm enables the transfer of semantic priors and cross-modal knowledge that are absent in vision-only approaches, potentially providing robustness under distribution shifts or limited training data.

CLIP has likely acquired sufficient prior knowledge to generalize to ``any" domain, which is a significant factor to consider; the main challenge lies in its suboptimal image-text alignment for certain tasks.
If so, it would be reasonable to ask such a question: can CLIP adapt to domains where its general-purpose natural language supervision is insufficient?
Before dealing with this broad research question, we first present a more specific inquiry:
\emph{Can CLIP understand depth?}

In this paper, we demonstrate that CLIP~\cite{radford2021learning} can be effectively adapted for monocular depth estimation, despite its suboptimal vision-language alignment during pretraining, without requiring fine-tuning. 
We propose CLIP2Depth, a framework that corrects the CLIP prior for depth estimation by training a single non-human language embedding matrix, called ``\emph{mirror}", along with a lightweight deconvolutional Transformer decoder~\cite{Luddecke_2022_CVPR}. 
An overview of the differences between previous human-language-based approaches and our proposed non-human language supervision is illustrated in \figref{fig:teaser}.
The concept of \emph{mirror} is inspired by prior works, notably the learnable depth prompts introduced by Auty~\etal~\cite{auty2023learning} and Hu~\etal~\cite{hu2023learning}, but it fundamentally differs in its non-human linguistic formulation and its role as a dynamic, learnable prompt rather than a static, fixed language prompt.
Instead of using pretrained subword embeddings, \emph{mirror} serves as a non-human language query input, which is passed through the frozen CLIP text encoder during training to distill its pretrained semantic prior in a non-human manner. 
Our ablation studies in Section~\ref{sec:ablation} clearly demonstrate the contribution of \emph{mirror} in improving depth estimation, particularly in identifying semantic objects such as driving cars, pedestrians, and transparent windows, where semantic cues are critical.
Our contributions are summarized as follows:
\begin{enumerate}{\leftmargin=4mm \itemindent=0mm \itemsep=0mm}
    \item 
    We demonstrate that CLIP can be effectively generalized to monocular dense depth estimation using its existing prior knowledge, all without fine-tuning, validating our title \emph{``CLIP Can Understand Depth."}
    \item 
    We show that a single query vector is sufficient to extract a single type of pixel-wise feature, eliminating the need for human-designed prompt sets or token binning.
    \item 
    Our model outperforms all previous CLIP-based depth estimation methods on NYU Depth v2~\cite{silberman2012indoor} and KITTI~\cite{geiger2013vision}.  
    Furthermore, it performs competitively with task-specific vision models while fully preserving the task-agnostic characteristic of the original CLIP. 
\end{enumerate}

\section{Related Works}
\subsection{CLIP-based Depth Estimation}
Early vision-language models typically handled images and text independently, as seen in works like Quattoni~\etal~\cite{quattoni2007learning} and Yuan~\etal~\cite{yuan2021multimodal}.
Although these works explored multimodal representation learning, they did not fully integrate vision and language into a shared embedding space. 
The introduction of CLIP~\cite{radford2021learning} marked a significant improvement in vision-language modeling by jointly training image and text encoders on large-scale datasets.
Later studies~\cite{gao2023clip,zhang2022can} further reinforced CLIP’s adaptability by demonstrating its effectiveness in few-shot domain adaptation.

Despite CLIP’s success in numerous downstream tasks, its application to specialized domains such as monocular depth estimation, where its pre-trained vision-language alignment is not naturally suited, remains relatively underexplored.
Similar domain-specific adaptation challenges have also been reported in microbial segmentation~\cite{elmessery2024semantic}, broiler weight estimation~\cite{shams2025automated}, and medical image classification~\cite{eliwa2024secure}.
Initial efforts by Zhang~\etal~\cite{zhang2022can} employed semantic text tokens describing distance extents to match image patches to depth bins in a zero-shot manner, aiming to validate whether CLIP’s pre-trained prior could capture depth cues without additional training. 
Auty~\etal~\cite{auty2023learning} extended this approach by introducing learnable distance tokens with a fixed human-language prefix, 
while Hu~\etal~\cite{hu2023learning} studied the few-shot depth estimation by using a learnable prefix prompt and fixed distance tokens. 
Their research is noteworthy for demonstrating that human-language-based distance tokens used to adapt CLIP's vision-language alignment for depth estimation have clear limitations. 
However, they all share a critical limitation: 
they use separate, predefined bin values to map image-text similarities to depth values, or they rely on natural language tokens to guide trainable parts of the prompt.

More recently, CaBins~\cite{Son_2024_CVPR} and WorDepth~\cite{zeng2024wordepth} have achieved notable performance improvements by leveraging CLIP’s text encoder prior. However, unlike the earlier approaches, they partially unfreeze CLIP during training~\cite{Son_2024_CVPR} or introduce a large number of additional trainable parameters via separate neural modules~\cite{Son_2024_CVPR, zeng2024wordepth}.
HoCLIP~\cite{zou2025high}, in particular, fine-tunes the CLIP backbone for monocular depth estimation. It enhances multi-modal fusion through high-order image feature modeling and learnable prompts and further refines depth predictions with an efficient multi-scale attention decoder and a vertical discriminator. 
These methods reflect a common trend: relying on CLIP fine-tuning combined with large auxiliary modules to improve performance on depth estimation tasks.

In contrast, our approach avoids both CLIP fine-tuning and the addition of large trainable modules. 
Instead, we propose a lightweight design based on a single, non-human language-learnable token that dynamically adjusts CLIP's vision-language alignment for depth estimation. 
Furthermore, we demonstrate that dense depth prediction can be achieved with a minimal deconvolutional decoder, effectively simplifying the model architecture while preserving performance.

While our focus is on CLIP-based approaches, it is worth noting that unimodal vision-only methods such as ZoeDepth~\cite{bhat2023zoedepth} and DepthAnything v2~\cite{yang2024depth} 
have recently achieved state-of-the-art results on standard benchmarks by scaling up very large backbones (\eg, BeiT-L, ViT-L) with hundreds of millions of parameters. 
This complementary line of research highlights a different trade-off: maximizing raw accuracy through scale, whereas our work emphasizes lightweight adaptation of frozen vision-language priors.

\subsection{Prompt Learning}
Prompt learning aims to optimize prompts for improved downstream performance while preserving model generalizability.
Early efforts focused on manually crafting prompts for specific tasks~\cite{radford2019language,brown2020language},
while later research explored automated prompt generation methods~\cite{shin2020autoprompt,wang2022learning}.
Although efficient, these automated methods sometimes introduced biases originating from human design choices.
CoOp~\cite{zhou2022learning} advanced the field by employing learnable vectors as input prompts, replacing discrete human language with continuous representations to enhance few-shot classification across various datasets.
Building on this foundation, CoCoOp~\cite{zhou2022conditional} introduced learnable meta-tokens to further improve generalizability.

In this work, we adapt CLIP for monocular depth estimation, a task that remains underexplored by conventional prompt learning approaches. 
We propose using a single learnable token to encode a universal depth query, enabling CLIP to perform dense depth estimation without relying on human-language prompts or preset bin values.
Our approach efficiently aligns CLIP's prior knowledge with task-specific requirements without requiring fine-tuning or adding large trainable modules, thereby advancing prompt learning research by extending its application from classification to regression tasks.

\section{Methodology}
\label{sec:method}

\begin{figure*}[!t]
\centering
\includegraphics[width=1.0\linewidth]{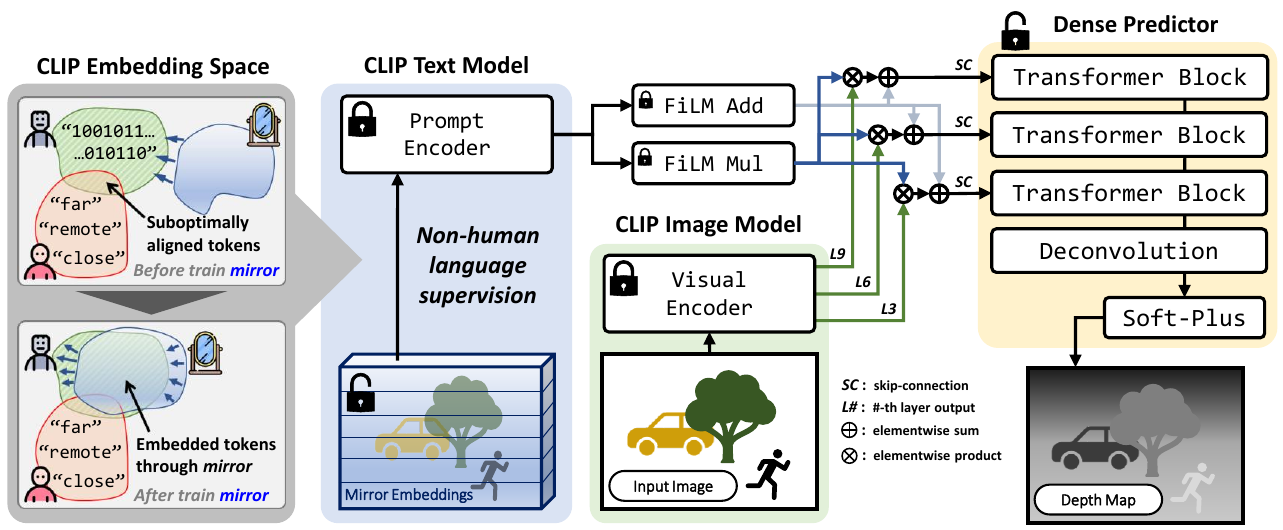}    
\caption{
  \textbf{Visualization of the proposed CLIP2Depth framework for depth estimation utilizing the \emph{mirror}.}
  In the detailed view on the right, open locks represent trainable components, while closed locks indicate frozen components.
  The \emph{mirror} is used as input to the CLIP text encoder, replacing human-language prompts that are suboptimal for depth estimation.
}
\label{fig:overview}
\end{figure*}

\begin{table}[b]
\centering 
\renewcommand{\arraystretch}{1.3}
\caption{Detailed architecture of CLIP2Depth model.}
\label{tab:architecture}
\resizebox{\textwidth}{!}{
    \begin{tabular}{llccccc}
    \toprule
    \textbf{Model} & \textbf{Module} & \textbf{Layer} & \textbf{Width} & \textbf{Head} & \textbf{FC layer} & \textbf{CNN} \\
    \midrule
    CLIP Image Model~\cite{radford2021learning} & Transformer Block & 12 & 768 & 12 & $768 \rightarrow 3072 \rightarrow 768$ & conv2d, $k{=}16$, $s{=}16$, $p{=}0$, $768$ \\
    CLIP Text Model~\cite{radford2021learning} & Transformer Block & 12 & 512 & 8 & $512 \rightarrow 2048 \rightarrow 512$ & \textendash \\
    \midrule
    \multirow{4}{*}{Dense predictor~\cite{Luddecke_2022_CVPR}} 
    & Transformer Block & 3 & 64 & 4 & $64 \rightarrow 2048 \rightarrow 64$ & \textendash \\
    & FiLM~\cite{perez2018film} & 2 & \textendash & \textendash & $512 \rightarrow 64$ & \textendash \\
    & Projection & 3 & \textendash & \textendash & $768 \rightarrow 64$ & \textendash \\
    & Deconvolution & \textendash & \textendash & \textendash & \textendash & 
    \begin{tabular}[c]{@{}l@{}}
        conv2d, $k{=}3$, $s{=}3$, $p{=}1$, $64$ \\ 
        convTrans2d, $k{=}4$, $s{=}4$, $p{=}0$, $32$ \\ 
        convTrans2d, $k{=}4$, $s{=}4$, $p{=}0$, $1$ 
    \end{tabular} \\
    \bottomrule
    \end{tabular}
}
\end{table}
We present the key components of CLIP2Depth, as illustrated in \figref{fig:overview}, highlighting how the non-human language \emph{mirror} embeddings are used within the frozen CLIP model~\cite{radford2021learning} for depth estimation.
The detailed architecture is summarized in Table~\ref{tab:architecture}, including CLIP image, text model and dense predictor.

\subsection{Non-human Language Prompt Learning} 
As shown in the experiments by Zhang~\etal~\cite{zhang2022can}, CLIP struggles to \emph{\textbf{consistently}} correlate images with depth-related prompts, highlighting the limitations of its token embeddings and text encoder in capturing depth cues.
One straightforward solution is to fine-tune CLIP for depth estimation while preserving its pre-trained knowledge from contrastive learning. 
However, this approach involves significant costs due to the scarcity of depth-text training examples. 
To address this, we propose a more cost-effective and scalable method: 
we freeze CLIP’s text encoder and distill its pre-trained semantic prior into a compact learnable embedding matrix called ``\emph{mirror}''.
The concept of \emph{mirror} is inspired by prior works~\cite{zhang2022can,auty2023learning,hu2023learning} that attempt to correct CLIP's prior through prompt learning, but it fundamentally differs in three key aspects:

\begin{itemize}
    \item \emph{Mirror} is a non-human language embedding matrix that fully eliminates the use of natural language token embeddings trained with CLIP.
    \item \emph{Mirror} is encoded into a single query vector, regardless of the number of its constituent embeddings.
    \item \emph{Mirror} maintains its unique role in depth estimation, capturing semantic objects in the scene even when jointly trained with larger modules.
\end{itemize}

When \emph{mirror} is formulated as an $(s, d)$ embedding matrix consisting of $s$ learnable vectors of $d$ dimensions, 
it is transformed into a single $(1, d)$ query embedding by passing through the frozen CLIP text encoder, 
in contrast to related works~\cite{zhang2022can,auty2023learning,hu2023learning} that generate $s$ query embeddings for $s$ learnable vectors.
This $(1, d)$ query embedding is then used to modulate~\cite{perez2018film} the image representation for depth estimation, 
whereas previous approaches aggregate $s$ embeddings (or their similarity scores) into a single depth value.
Although we do not provide explicit empirical comparisons, our method avoids the over-smoothing effect caused by aggregating multiple embeddings and reduces computational overhead, particularly as the number of learnable embeddings increases.
While it may seem odd to condition all images on the same query vector, 
we argue that a single query such as ``\emph{How far is this location from the camera?}" is sufficient for depth estimation, regardless of the image's specific content.
After training with the frozen CLIP encoders for depth estimation, 
\emph{mirror} successfully learns to capture semantic objects such as driving cars, windows, and pedestrians, 
where understanding the semantic context is crucial for accurate identification (see \figref{fig:ablation_mirror}).

\subsection{Minimal Decoder for Dense Prediction}
The image encoder of CLIP encodes an input image into multiple patch embeddings, allowing the computation of similarity scores between each patch and the text embeddings without requiring any additional decoder layers.
However, the spatiotemporal information that is lost at this point makes it difficult to accurately capture depth information.
To address this, we introduce a deconvolutional decoder layer to enable dense prediction.
Rather than designing a new decoder module from scratch, 
we investigate a proven lightweight decoder, specifically using CLIPSeg~\cite{Luddecke_2022_CVPR}, as it is one of the most lightweight and validated solutions among existing methods, ensuring that the model does not rely heavily on the decoder for learning depth representations.

We fine-tune a pre-trained CLIPSeg decoder with the \emph{mirror} for depth estimation through the following forward pass:
\[
\mathrm{Pred}(x) = D(E_{\text{visual}}(x), E_{\text{prompt}}(\mathrm{concat(BOS, M, EOS)})^{*})
\]
where $x \in \mathbb{R}^{352\times352}$ denotes a resized input image, and $\mathrm{M} \in \mathbb{R}^{64\times512}$ represents the \emph{mirror} consisting of 64 learnable 512-dimensional latent vectors.
To form the input to the CLIP text encoder, we concatenate the special CLIP token embeddings BOS and EOS to the beginning and end of $\mathrm{M}$, respectively.

Here, $D$ denotes the CLIPSeg decoder, and $E_{\text{visual}}$ and $E_{\text{prompt}}$ refer to the CLIP image encoder and text encoder, respectively.
The superscript asterisk `*' denotes the eos-pooling operation, where $E_{\text{prompt}}(\mathrm{concat(BOS, M, EOS)})^{*}$ yields a projected \texttt{<|eos|>} embedding with 512 dimensions.
This conditional embedding is passed to the decoder $D$ and used to modulate the image representation $E_{\text{visual}}(x) \in \mathbb{R}^{484\times768}$ via two FiLM~\cite{perez2018film} layers.
The number of patch embeddings, 484, is obtained by dividing a $352\times352$ image into $16\times16$ patches, following the ViT-B/16 configuration, consistent with the original CLIPSeg setup for 2D semantic segmentation.

\algrenewcommand\algorithmiccomment[1]{\hfill \textcolor{teal}{\(\triangleright\) #1}}

\noindent
\begin{minipage}{\textwidth}
\vspace{0pt}
    \begin{algorithm}[H]
    \footnotesize
    \caption{\textbf{Training of CLIP2Depth}}
    \textbf{Input:} A batch of images $X \in \mathbb{R}^{B \times C \times H \times W}$, corresponding depth maps $Y \in \mathbb{R}^{B \times H \times W}$ \\
    \textbf{Output:} Updated parameters $\theta_{\text{mirror}}, \theta_D$ \\
    \textbf{Initialization:} Pre-trained CLIP image encoder $E_{\text{visual}}$, pre-trained CLIP text encoder $E_{\text{prompt}}$, pre-trained CLIPSeg decoder $D$, special token embeddings $e_{\text{BOS}}, e_{\text{EOS}}$, learnable embeddings $mirror$
        \begin{algorithmic}[1]
        \setstretch{1.2}
        \Function{CLIP2Depth\_Training}{$X, Y$}
            \State $B \gets$ batch size of $X$
            \State $h \gets E_{\text{visual}}^{\{3,6,9\}}(X)$ \Comment{select intermediate hidden states}
            \State $e^{B}_{\text{BOS}}, e^{B}_{\text{EOS}}, e^{B}_{\text{mirror}} \gets \texttt{repeat}(e_{\text{BOS}}, B), \texttt{repeat}(e_{\text{EOS}}, B), \texttt{repeat}(mirror, B)$
               \Comment{replicate for batch}
            \State $e_{\text{concat}} \gets \texttt{concat}(e^{B}_{\text{BOS}}, e^{B}_{\text{mirror}}, e_{\text{EOS}}^{B})$ 
            \State $z_{\text{cond}} \gets E_{\text{prompt}}(e_{\text{concat}})$
            \State $logits \gets D(\text{hidden\_states}=h$, \Comment{decode}\\
                \hspace*{6em} $\text{conditional\_embeddings} = \text{eos\_pooling}(z_{\text{cond}}))$ 
            \State $\text{Pred}(X) \gets \texttt{softplus}(logits)$ \Comment{predict depth map}
            \State $\hat{d} \gets \texttt{interpolate}(\text{Pred}(X), \text{target\_size}=Y.\text{shape}, \text{interpolation}=\text{bilinear})$
            \State $\mathcal{L}_{\text{ScaleInvariant}} \gets \texttt{ScaleInvariantLoss}(\hat{d}, Y)$ 
                \Comment{compute scale-invariant depth loss}
            \State $(\theta_{\text{mirror}}, \theta_D) \gets \texttt{Optimizer\_step}(\mathcal{L}_{\text{ScaleInvariant}}, \{\theta_{\text{mirror}}, \theta_D\})$ \hfill 
                \Comment{update $mirror$ and decoder}
        \EndFunction
        \end{algorithmic}
    \label{algo:clip2depth_train}
    \end{algorithm}
    \vspace{-1.5em} 
    \begin{algorithm}[H]
    \footnotesize
    \caption{\textbf{Inference of CLIP2Depth}}
    \textbf{Input:} Image $X \in \mathbb{R}^{C \times H \times W}$ \\
    \textbf{Output:} Predicted depth map $\hat{d}$ \\
    \textbf{Initialization:} Pre-trained CLIP image encoder $E_{\text{visual}}$, pre-trained CLIP text encoder $E_{\text{prompt}}$, fine-tuned CLIPSeg decoder $D$, special token embeddings $e_{\text{BOS}}, e_{\text{EOS}}$, learned embeddings $mirror$
        \begin{algorithmic}[1]
        \setstretch{1.2}
        \Function{CLIP2Depth\_Inference}{$X$}
            \State $h \gets E_{\text{visual}}^{\{3,6,9\}}(X)$ \hfill 
                \Comment{select intermediate hidden states}
            \State $e_{\text{concat}} \gets \texttt{concat}(e_{\text{BOS}}, e_{\text{mirror}}, e_{\text{EOS}})$ \hfill 
            \State $z_{\text{cond}} \gets E_{\text{prompt}}(e_{\text{concat}})$
            \State $logits \gets D(\text{hidden\_states}=h,$ \hfill 
                \Comment{decode} \\
            \hspace*{6.5em} $\text{conditional\_embeddings} = \text{eos\_pooling}(z_{\text{cond}}))$
            \State $\text{Pred}(X) \gets \texttt{softplus}(logits)$ \hfill 
                \Comment{predict depth map}
            \State $\hat{d} \gets \texttt{interpolate}(\text{Pred}(X), \text{target\_size}=X.\text{shape}, \text{interpolation}=\text{bilinear})$
            \State \textbf{return} $\hat{d}$
        \EndFunction
        \end{algorithmic}
    \label{algo:clip2depth_inference}
    \end{algorithm}
\end{minipage}

\vspace{1em} 

\paragraph{\textbf{Implementation details}}
Besides freezing the CLIP encoders, we also freeze the two feed-forward networks within the FiLM~\cite{perez2018film} blocks, which slightly improves the convergence of the \emph{mirror} and enables more consistent semantic cue extraction across various types of input scenes.
The remaining parts of the decoder $D$ are jointly trained with the \emph{mirror} $\mathrm{M}$, which is initialized from a normal distribution with mean $\mu=0.0$ and standard deviation $\sigma=0.02$.
During fine-tuning, we minimize the modified scale-invariant loss from AdaBins~\cite{bhat2021adabins}, using hyperparameters $\lambda=0.85$ and $\alpha=10$, under a supervised setting using annotations of ground-truth depth.

To complement our architectural description, we present Algorithm~\ref{algo:clip2depth_train} and Algorithm~\ref{algo:clip2depth_inference}, which describe the procedures of CLIP2Depth during training and inference, respectively.
Algorithm~\ref{algo:clip2depth_train} outlines the full training procedure: extracting intermediate visual features from a frozen CLIP image encoder, conditioning the CLIPSeg decoder using the learnable \emph{mirror} embedding through a text encoder, and computing depth predictions with a scale-invariant loss. Gradients are applied only to the decoder and the \emph{mirror} embedding, in line with our lightweight design principles.
Algorithm~\ref{algo:clip2depth_inference} illustrates the inference-time forward pass, where the same \emph{mirror}-conditioned depth prediction is performed using the pre-trained encoders and fixed decoder, but without any parameter updates.

\section{Experiments}
\label{sec:experiment}

\subsection{Experimental Setup} 
\subsubsection{Datasets}
We adopt the NYU Depth v2~\cite{silberman2012indoor} and KITTI~\cite{geiger2013vision} datasets, two standard benchmarks for monocular depth estimation, and follow the official splits provided by AdaBins~\cite{bhat2021adabins}.
The key characteristics of both datasets are summarized in Table~\ref{tab:dataset_summary}.    
For zero-shot generalization, we additionally use the SUN RGB-D dataset~\cite{song2015sun} as a target domain for evaluation.

The \textbf{NYU Depth v2} dataset contains RGB-D images collected from various indoor scenes such as living rooms, bedrooms, and offices. It includes over 464 unique scenes, providing a rich variety of layouts, objects, and lighting conditions. Depth maps are densely annotated using a Microsoft Kinect sensor, providing aligned RGB and depth images at a resolution of 640 $\times$ 480. We follow the official split with 24,231 training images and 654 test images. 
During training, we apply the Eigen cropping strategy~\cite{eigen2014depth}, followed by resizing all images to 352 $\times$ 352. 
For evaluation, all images are resized to 352 $\times$ 352 without cropping.

The \textbf{KITTI} dataset focuses on outdoor driving scenes in urban, rural, and highway settings. It provides sparse depth maps generated by projecting LiDAR point clouds onto RGB images captured by a stereo camera, with an original resolution of 1,241 $\times$ 376. 
The dataset contains 61 distinct driving sequences, covering diverse traffic scenarios, comprising 42,746 frames with synchronized sensor data, including LiDAR, stereo RGB cameras, and GPS/IMU measurements. 
We adopt the Eigen split~\cite{eigen2014depth}, using 23,157 images for training and 696 for testing. 
For both training and evaluation, we follow the cropping strategy proposed by Garg~\etal~\cite{garg2016unsupervised} and resize all images to 352 $\times$ 352. 
Additionally, for spatial continuity analysis~(Section~\ref{sec:experiment_continuous}), we also utilize the KITTI object detection dataset~\cite{Zhu_2019_ICCV}.

In addition, we use the SUN RGB-D dataset~\cite{song2015sun} for zero-shot generalization~(Section~\ref{sec:zero-shot}). 
The dataset provides RGB-D images at a resolution of 730 $\times$ 530, captured with multiple sensors across 47 indoor scene categories such as kitchens, offices, and bedrooms. 
In our experiments, we use only the official test split (5,050 images) to evaluate zero-shot generalization from models trained on NYU Depth v2.

\begin{table}[t]
\renewcommand*{\arraystretch}{1.5}
\centering
\captionsetup{font=normalsize}
\caption{Comparison of dataset characteristics and the number of images in each split.}
\small
\label{tab:dataset_summary}
    \begin{tabular}{lccccc}
    \toprule
    \textbf{Dataset} & \textbf{Domain} & \textbf{Total} & \textbf{Train} & \textbf{Test} & \textbf{Resolution} \\
    \midrule
    NYU Depth v2~\cite{silberman2012indoor} & Indoor (RGB-D) & 24,885 & 24,231 & 654 & 640  $\times$ 480 \\
    KITTI~\cite{geiger2013vision} & Outdoor (RGB + LiDAR) & 23,853 & 23,157 & 696 & 1,241  $\times$ 376 \\
    \bottomrule
    \end{tabular}
\end{table}
    
\subsubsection{Training Details}
We train all models for 25 epochs with a batch size of 32, using identical training settings for both the NYU Depth v2 and KITTI datasets.
We employ the AdamW optimizer~\cite{loshchilov2018decoupled} with a weight decay of $\lambda = 0.01$, and adopt cosine annealing~\cite{loshchilov2017sgdr} for learning rate scheduling, starting from an initial learning rate of 0.003.
Training is performed on a single NVIDIA RTX 3090 GPU with mixed-precision computation and an i9-10900X CPU. All experiments are conducted using Python 3.9, PyTorch 2.0.1, and CUDA 11.7.

\subsubsection{Evaluation Metrics}
We evaluate the performance of our method using standard depth estimation metrics introduced by Eigen~\etal~\cite{eigen2014depth}: 
the absolute relative error (Abs Rel), defined as $\frac{1}{T} \sum_{i \in T} \frac{|d_i - d_i^{gt}|}{d_i^{gt}}$;
the squared relative error (Sq Rel), $\frac{1}{T} \sum_{i \in T} \frac{(d_i - d_i^{gt})^2}{d_i^{gt}}$;
the root mean squared error (RMSE), $\sqrt{\frac{1}{T} \sum_{i \in T} (d_i - d_i^{gt})^2}$;
and the log$_{10}$ error, $\frac{1}{T} \sum_{i \in T} |\log_{10}(d_i) - \log_{10}(d_i^{gt})|$.
Additionally, we report the threshold accuracy $\delta_n$, which measures the percentage of predictions $d_i$ satisfying $\max\left(\frac{d_i}{d_i^{gt}}, \frac{d_i^{gt}}{d_i}\right) < \text{threshold}$, where the threshold is set to $1.25^n$ for $n \in \{1, 2, 3\}$.

\begin{figure*}[]
    \centering
    \includegraphics[width=1.0\linewidth]{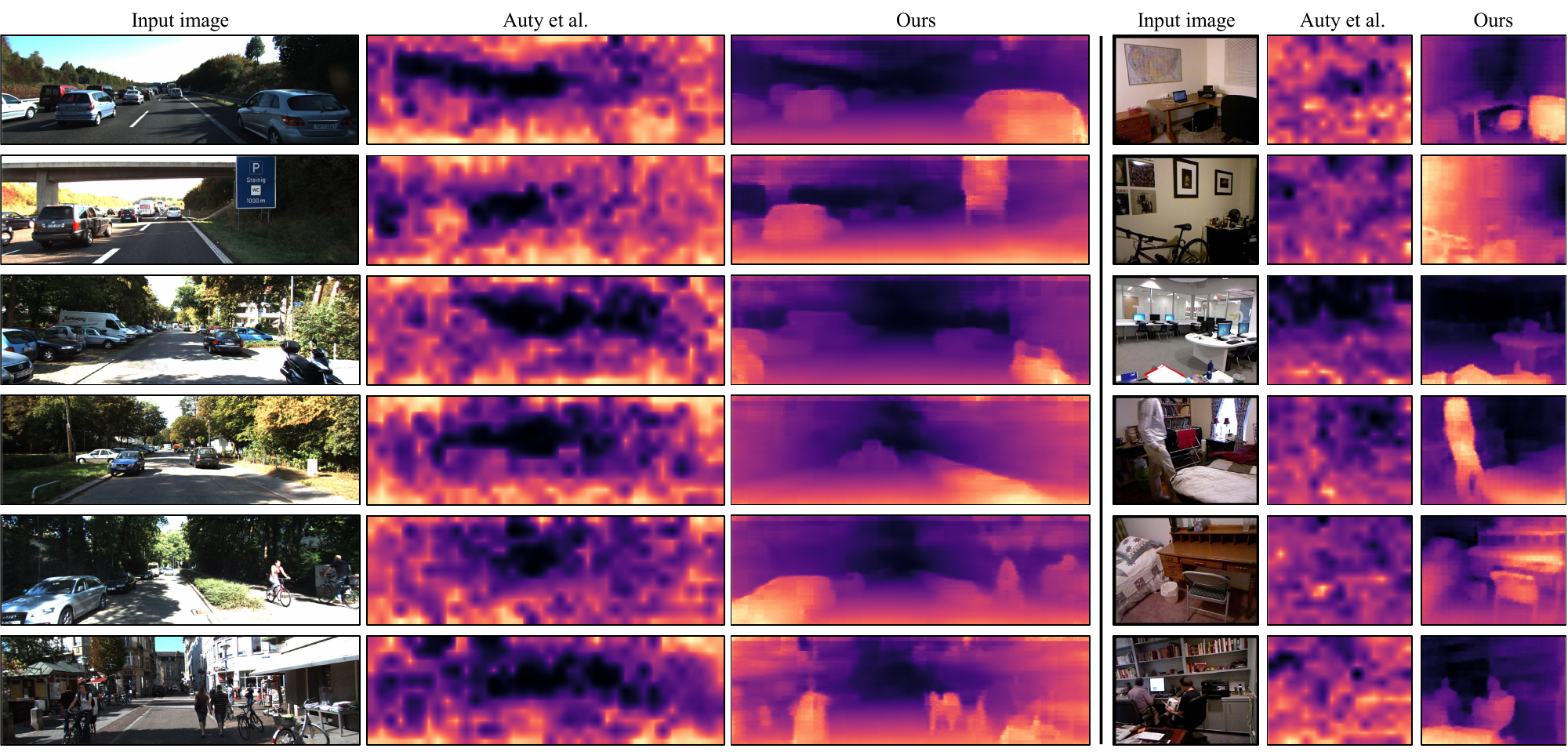}
    \captionsetup{font=normalsize}
    \caption{
    Qualitative comparison of monocular depth estimation between the method proposed by Auty~\etal~\cite{auty2023learning} and ours on the KITTI~\cite{geiger2013vision} dataset (left) and the NYU Depth v2~\cite{silberman2012indoor} dataset (right). All results are absolute and inverse depth using the \emph{magma} colormap.}
    \label{fig:main}
\end{figure*}

\begin{table*}[t!]
  \small
  \renewcommand*{\arraystretch}{1.8}
    \captionsetup{font=normalsize}
    \caption{
      Comparison of depth estimation performance on the NYU Depth v2~\cite{silberman2012indoor} dataset with representative unimodal vision models and previous CLIP-based approaches.
      ZoeDepth~\cite{bhat2023zoedepth} results are taken from ZeroDepth~\cite{guizilini2023towards}.
      CaBins~\cite{Son_2024_CVPR} and HoCLIP~\cite{zou2025high} are listed separately since they fine-tune at least one of the CLIP encoders during training.
      `\textendash' indicates that the metric is not reported in the original paper~\cite{saxena2008make3d}.
      \textbf{Bold} indicates the best performance.
    }
    \label{tab:nyu}
  \begin{tabularx}{\textwidth}{Xcccccc}
    \Xhline{\arrayrulewidth}
    \hline
    \textbf{Model} & $\bf Abs\,Rel\downarrow$ & $\bf RMSE\downarrow$ & $\mathrm{\bf{log_{10}}}\downarrow$ & $\bf \delta^{1}\uparrow$ & $\bf \delta^{2}\uparrow$ & $\bf \delta^{3}\uparrow$ \\
    \hline
    Make3D~\cite{saxena2008make3d} & 0.349            & 1.214           & \textendash              & 0.447          & 0.745          & 0.897 \\
    DORN~\cite{fu2018deep} & 0.115            & 0.509           & 0.051           & 0.828          & 0.965          & 0.992 \\
    ASTransformer~\cite{chang2021transformer} & 0.103            & 0.374           & 0.044           & 0.902          & 0.985          & 0.997 \\
    DepthFormer~\cite{li2023depthformer} & 0.096            & 0.339           & 0.041           & 0.921          & 0.989          & 0.998 \\
    NeWCRFs~\cite{yuan2022neural} & 0.095   & 0.334  & 0.041  & 0.922 & 0.992 & 0.998 \\
    ZoeDepth~\cite{bhat2023zoedepth} & 0.077 & 0.277 & 0.033 & 0.953 & 0.995 & 0.999 \\
    DepthAnything~v2~\cite{yang2024depth} & \textbf{0.056} & \textbf{0.206} & \textbf{0.024} & \textbf{0.984} & \textbf{0.998} & \textbf{1.000} \\
    \hline
    DepthCLIP~\cite{zhang2022can} & 0.388            & 1.167           & 0.156           & 0.394          & 0.683          & 0.851 \\
    Hu~\etal~\cite{hu2023learning}    & 0.347            & 1.049           & 0.140           & 0.428          & 0.732          & 0.898 \\
    Auty~\etal~\cite{auty2023learning}  & 0.319            & 0.970           & 0.128           & 0.465          & 0.776          & 0.922 \\
    \rowcolor[HTML]{EFEFEF}
    \textbf{CLIP2Depth} & \textbf{0.100}   & \textbf{0.379}  & \textbf{0.042}  & \textbf{0.900} & \textbf{0.983} & \textbf{0.996} \\
    \hline
    CaBins~\cite{Son_2024_CVPR} & 0.120   & 0.041  & 0.050  & 0.866 & 0.978 & 0.996 \\
    HoCLIP~\cite{zou2025high} & 0.214 & 0.702 & 0.085 & 0.668 & 0.914 & 0.978 \\
    \Xhline{\arrayrulewidth}
  \end{tabularx}  
\end{table*}

\begin{table*}[t]
  \footnotesize
  \renewcommand*{\arraystretch}{1.8}
    \captionsetup{font=normalsize}
    \caption{
      Comparison of depth estimation performance on the KITTI~\cite{geiger2013vision} dataset with representative unimodal vision models and previous CLIP-based approaches.
      ZoeDepth~\cite{bhat2023zoedepth} results are taken from ZeroDepth~\cite{guizilini2023towards}.
      CaBins~\cite{Son_2024_CVPR} and HoCLIP~\cite{zou2025high} are listed separately since they fine-tune at least one of the CLIP encoders during training.
      `\textendash' indicates that the metric is not reported in the original paper~\cite{chang2021transformer, auty2023learning}.
      }
  \begin{tabularx}{\textwidth}{Xccccccc}
    \Xhline{\arrayrulewidth}
    \hline
    \textbf{Model} & $\bf Abs\,Rel\downarrow$ & $\bf Sq Rel\downarrow$ & $\bf RMSE\downarrow$ & $\mathrm{\bf{log_{10}}}\downarrow$ & $\bf \delta^{1}\uparrow$ & $\bf \delta^{2}\uparrow$ & $\bf \delta^{3}\uparrow$ \\
    \hline
    DORN~\cite{fu2018deep} & 0.072            & 0.307           & 2.727          & 0.120          & 0.932          & 0.984    & 0.994 \\
    ASTransformer~\cite{chang2021transformer}  & 0.103            & \textendash           & \textendash           & 0.374          & 0.963          & 0.995   & 0.999 \\
    DepthFormer~\cite{li2023depthformer}  & 0.052            & 0.158           & 2.143           & 0.079          & 0.975          & 0.997          & 0.999          \\
    NeWCRFs~\cite{yuan2022neural} & 0.052   & \textbf{0.155}  & 2.129  & 0.079 & 0.974 & 0.997 & 0.999 \\
    ZoeDepth~\cite{bhat2023zoedepth} & 0.057 & \textendash & 2.290 & \textendash & 0.967 & 0.995 & 0.999 \\
    DepthAnything~v2~\cite{yang2024depth} & \textbf{0.045} & \textendash  & \textbf{1.861} & \textbf{0.060} & \textbf{0.983} & \textbf{0.998} & \textbf{1.000} \\
    \hline
    DepthCLIP~\cite{zhang2022can} & 0.473            & 6.007           & 12.958          & 0.680          & 0.281          & 0.531 & 0.696 \\
    Hu~\etal~\cite{hu2023learning} & 0.384            & 4.661           & 12.290           & 0.632          & 0.312          & 0.569 & 0.739 \\
    Auty~\etal~\cite{auty2023learning} & 0.303            & \textendash           & 6.322           & 0.112          & 0.550          & 0.830 & 0.938 \\
    \rowcolor[HTML]{EFEFEF}
    \textbf{CLIP2Depth}  & \textbf{0.074}   & \textbf{0.303}  & \textbf{2.948}  & \textbf{0.032} & \textbf{0.938} & \textbf{0.990}  & \textbf{0.998}\\
    \hline
    CaBins~\cite{Son_2024_CVPR} & 0.057   & 0.186  & 2.322  & 0.025 & 0.964 & 0.995 & 0.999 \\
    HoCLIP~\cite{zou2025high} & 0.107 & 0.539 & 3.835 & 0.045 & 0.879 & 0.975 & 0.995 \\
    \Xhline{\arrayrulewidth}
  \end{tabularx}

    \label{tab:kitti}
\end{table*}

\begin{table}[t]
    \centering
    \small
    \renewcommand*{\arraystretch}{1.5}
    \captionsetup{font=normalsize}
    \caption{Comparison of the number of learnable parameters and computational complexity (GFLOPs) on the KITTI dataset.
    GFLOPs are measured with ptflops (MACs $\times$ 2) for a single forward pass at the input resolution used for evaluation.
    The values of DORN~\cite{fu2018deep}, NeWCRFs~\cite{yuan2022neural}, and HoCLIP~\cite{zou2025high} are taken from HoCLIP, while the parameter count of Hu~\etal~\cite{hu2023learning} method is taken from its paper. 
    Since Hu~\etal did not release official code and did not report GFLOPs, it is shown as `\textendash’.
    }
    \label{tab:params}
    \begin{tabular}{lrr}
    \toprule
    \textbf{Model} & \textbf{Learnable Params} & \textbf{GFLOPs} \\
    \midrule
    DORN~\cite{fu2018deep}             & 110,300k & 775.05 \\
    NeWCRFs~\cite{yuan2022neural}      & 270,440k & 230.61 \\
    ZoeDepth~\cite{bhat2023zoedepth}    & 345,000k & 538.60 \\    DepthAnything~v2~\cite{yang2024depth} & 335,300k & 1121.58 \\
    \midrule
    DepthCLIP~\cite{zhang2022can}      & 0        & 48.76  \\
    Hu~\etal~\cite{hu2023learning}     & 8k       & \textendash     \\
    Auty~\etal~\cite{auty2023learning} & 3k       & 107.36 \\
    \rowcolor[HTML]{EFEFEF}
    \textbf{CLIP2Depth}        & 1,095k   & 89.60  \\
    \midrule
    CaBins~\cite{Son_2024_CVPR}        & 143,154k & 378.14 \\
    HoCLIP~\cite{zou2025high}          & 295,000k & 944    \\
    \bottomrule
    \end{tabular}
\end{table}

\subsection{Monocular Depth Estimation}
In this section, we compare our main experimental results with selected previous state-of-the-art and recently proposed works using a frozen CLIP~\cite{radford2021learning} prior for depth estimation.
To ensure clarity, we first outline the baseline comparison criteria.
While our method adopts a larger backbone (CLIP ViT-B/16) and includes an additional learnable decoder layer compared to prior works~\cite{auty2023learning,hu2023learning,zhang2022can} based on CLIP ResNet-50, we argue that the comparison remains valid because the CLIP prior is inherently suboptimal for depth estimation and is kept entirely frozen during training, without any task-specific adaptation.
Although this architectural difference may make direct performance comparisons less equitable, our primary objective is not to surpass previous methods in raw accuracy under equal model capacity but rather to demonstrate the feasibility and effectiveness of correcting the frozen CLIP prior for depth estimation without fine-tuning. 
Therefore, we present the results without adjusting for differences in model capacity.

In the context of our study, the only requirement for a fair comparison among CLIP-based approaches is that the CLIP encoder must remain frozen during training.
Since CaBins~\cite{Son_2024_CVPR} and HoCLIP~\cite{zou2025high} fine-tune CLIP, we exclude them from our baseline comparisons.
Quantitative results on the NYU Depth v2 and KITTI datasets are presented in \tabref{tab:nyu} and \tabref{tab:kitti}, respectively, while qualitative comparisons illustrating visual performance are shown in \figref{fig:main}.

We first compare our method with recent studies that utilize the pre-trained, frozen CLIP model for monocular depth estimation~\cite{auty2023learning,hu2023learning,zhang2022can}. 
To the best of our knowledge, ours is the first vision-language approach for depth estimation that performs comparably to unimodal (vision-only) models without fine-tuning any components of the pre-trained CLIP.
While CLIP2Depth substantially improves upon prior CLIP-based baselines, it is also important to assess its performance compared to recent state-of-the-art vision-only approaches such as ZoeDepth~\cite{bhat2023zoedepth} (based on BeiT-L) and DepthAnything~v2~\cite{yang2024depth} (based on ViT-L). 
These large-scale models still achieve higher raw accuracy by leveraging very large backbones and hundreds of millions of learnable parameters. 
In comparison, although CLIP2Depth is slightly behind in absolute accuracy, it offers a favorable trade-off between accuracy and efficiency: our model requires only 1.1M learnable parameters, compared to 346M for ZoeDepth and 335M for DepthAnything~v2, as summarized in \tabref{tab:params}. 
This demonstrates that correcting a frozen CLIP prior can already deliver competitive quality without heavy task-specific adaptation, while also clarifying the position of vision-language models within the broader landscape of monocular depth estimation.

Another notable point is that all previous CLIP-based approaches~\cite{auty2023learning,hu2023learning,zhang2022can} consistently exhibit poorer performance on KITTI than on NYU Depth v2. 
Our method shows a similar trend, but with a significantly reduced performance gap, as shown in \tabref{tab:kitti}. This suggests that the improvement does not come from increased parameters or the deconvolutional decoder, but rather from a more fundamental enhancement or correction in depth understanding.

Additionally, we provide a detailed comparison of learnable parameters and computational complexity in \tabref{tab:params}.  
Compared to vision-based depth estimation models, our method achieves competitive performance with significantly fewer learnable parameters.  
Although our model uses more parameters than previous CLIP-based depth estimation methods, it consistently achieves superior performance.  
Notably, HoCLIP~\cite{zou2025high}, which fine-tunes the entire CLIP encoder, requires substantially more learnable parameters. 
In contrast, our method outperforms it by updating only the \emph{mirror} embedding and a lightweight decoder, highlighting the effectiveness of our minimal intervention strategy in leveraging CLIP's pre-trained knowledge for depth estimation.

\begin{figure*}[t]
  \centering
  \includegraphics[width=1.0\linewidth]{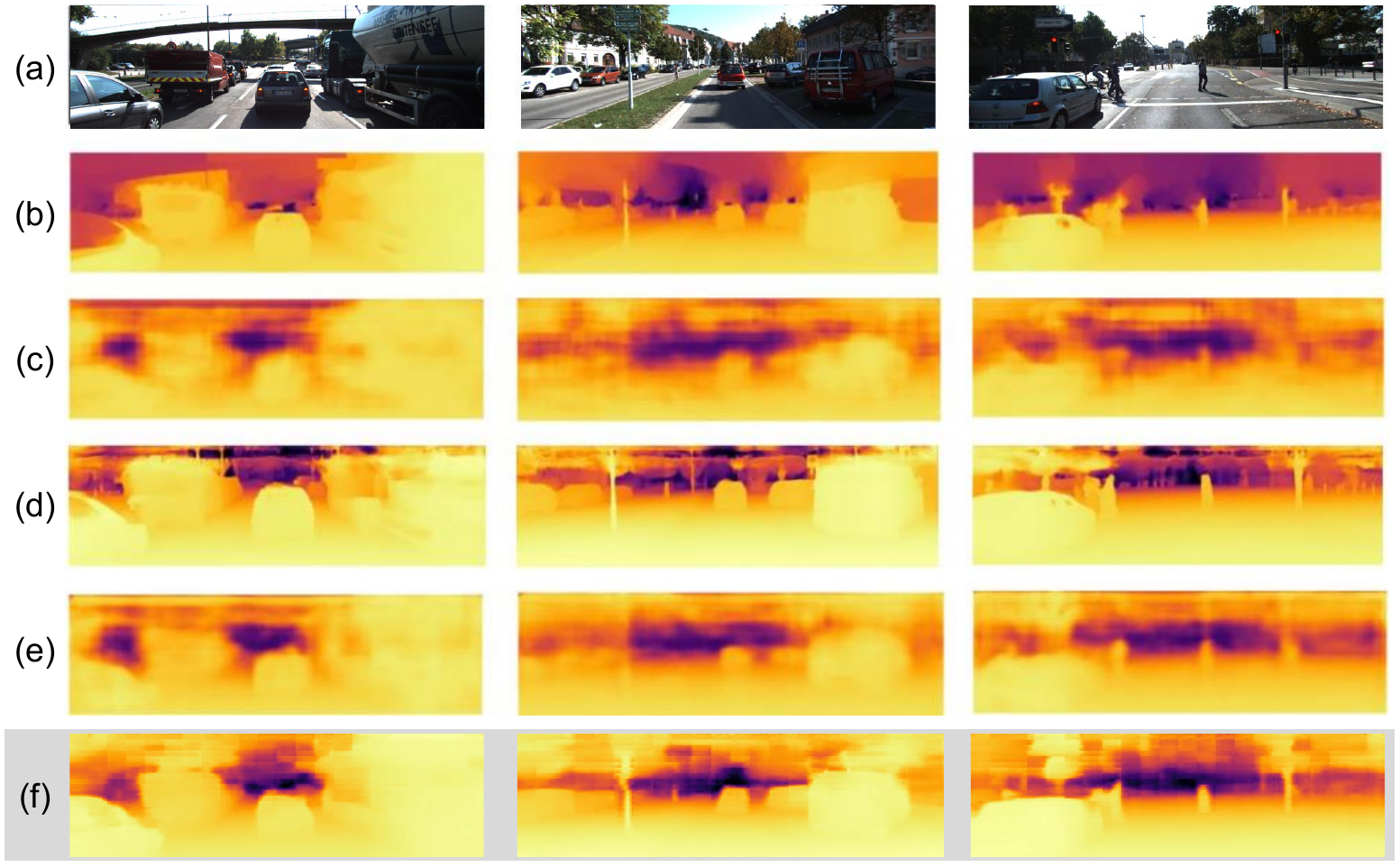} 
  \captionsetup{font=normalsize}
    \caption{
    Qualitative comparison on the KITTI dataset: (a) RGB input, (b) ground truth, (c) Auty~\etal~\cite{auty2023learning}, (d) BTS~\cite{lee2019big}, (e) HoCLIP~\cite{zou2025high}, and (f) our CLIP2Depth. 
    Rows (b–e) are taken from HoCLIP~\cite{zou2025high} for fair comparison. 
    While HoCLIP requires fine-tuning the entire CLIP encoder and 295M learnable parameters, CLIP2Depth achieves sharper object boundaries and more coherent structures with only 1.1M learnable parameters, highlighting the effectiveness of our lightweight strategy. Our result (f) is visualized as absolute linear depth using the \emph{inferno\_r} colormap.}
  \label{fig:qual_hoclip}
\end{figure*}

In addition, we provide a qualitative evaluation against representative methods, 
including Auty et al.~\cite{auty2023learning}, BTS~\cite{lee2019big}, and HoCLIP~\cite{zou2025high}, as shown in \figref{fig:qual_hoclip}.
Since HoCLIP has not released official code or pre-trained weights, we directly include its original results (b–e) from the paper and append our CLIP2Depth predictions (f) for side-by-side comparison. 
Notably, (d) corresponds to a conventional monocular depth estimation method, while (c), (e), and (f) are CLIP-based approaches.
As shown in the figure, CLIP2Depth surpasses prior CLIP-based methods by producing sharper object boundaries and more coherent scene structures, particularly around moving vehicles and road edges, while requiring nearly 300$\times$ fewer learnable parameters. 
Although its quality does not fully reach that of the strong supervised baseline BTS, 
CLIP2Depth achieves a favorable trade-off between accuracy and efficiency, 
demonstrating that frozen CLIP priors with minimal adaptation can already deliver competitive visual quality.

\begin{figure*}[]
  \centering
  \includegraphics[width=1.0\linewidth]{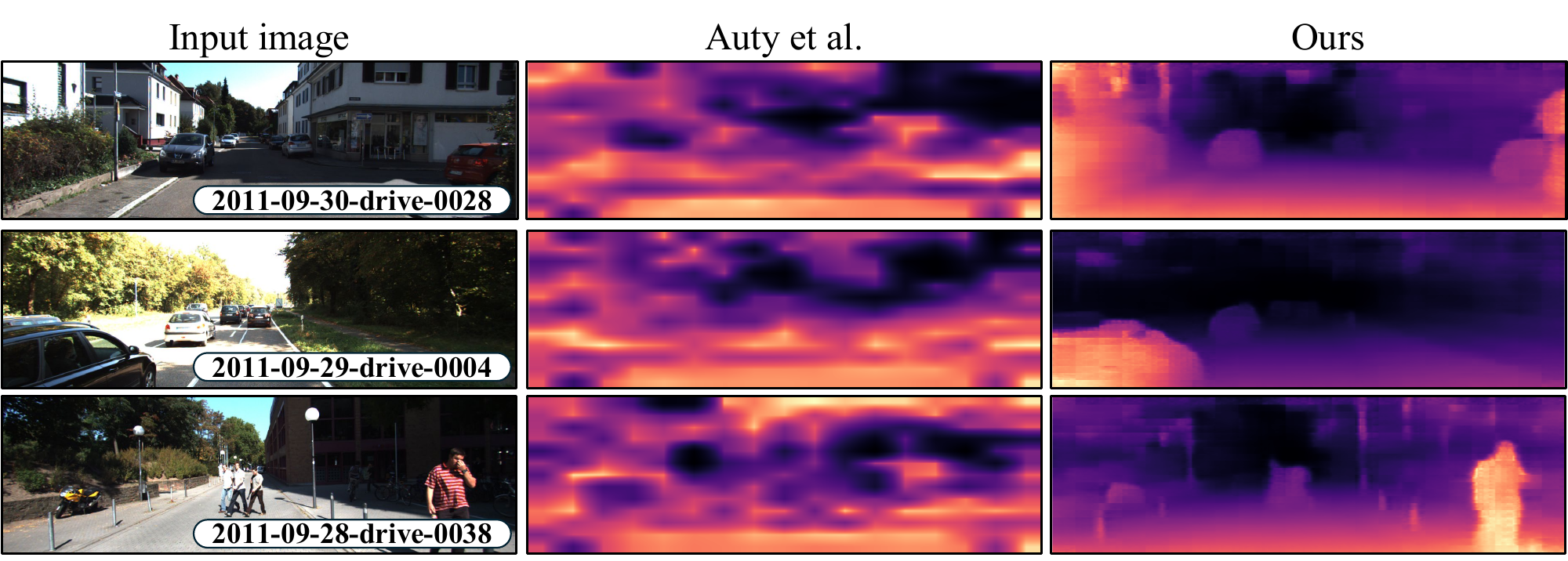} 
  \captionsetup{font=normalsize}
  \caption{
          Qualitative comparison on selected KITTI frames illustrating various motion scenarios.
          Each row presents a distinct scene type: 
          (top) static scene with potentially movable objects, 
          (middle) dynamic objects moving in the same direction as the camera, and 
          (bottom) dynamic entities approaching from the opposite direction.
          Our method demonstrates improved structural preservation and semantic alignment compared to the method of Auty~\etal~\cite{auty2023learning}, even under challenging motion patterns. 
          All results are absolute and inverse depth using the \emph{magma} colormap.
        }
  \label{fig:temporal_frames}
\end{figure*}
To further assess the robustness of our method under various motion scenarios, we visualize qualitative results on the KITTI dataset in \figref{fig:temporal_frames}. 
Each example features dynamic or potentially dynamic objects captured under different camera motion patterns, including static views, forward motion, and opposing directions. 
This figure compares ours with the method of Auty~\etal~\cite{auty2023learning}, demonstrating improved preservation of object structure and scene geometry more effectively, even in challenging dynamic scenes.

\begin{table}[t!]
\footnotesize
\centering
\renewcommand*{\arraystretch}{1.5}
\captionsetup{font=normalsize}
\caption{Zero-shot generalization results from NYU Depth v2 to SUN RGB-D (indoor scenes). 
The comparison includes representative vision-only models~\cite{bhat2021adabins, yuan2022neural, bhat2023zoedepth} and CLIP-based approaches~\cite{auty2023learning, zou2025high}.
The values of the method of Auty~\etal are taken from HoCLIP.}
\begin{tabular}{lcccccc}
    \toprule
    \textbf{Model} & $\bf Abs\,Rel\downarrow$ & $\bf RMSE\downarrow$ & $\mathrm{\bf{log_{10}}}\downarrow$ & $\bf \delta^{1}\uparrow$ & $\bf \delta^{2}\uparrow$ & $\bf \delta^{3}\uparrow$ \\
    \midrule
    AdaBins~\cite{bhat2021adabins}      & 0.159 & 0.476 & 0.068 & 0.771 & 0.944 & 0.983 \\
    NeWCRFs~\cite{yuan2022neural}       & 0.151 & 0.424 & 0.064 & 0.798 & 0.967 & 0.992 \\
    ZoeDepth~\cite{bhat2023zoedepth}    & \textbf{0.119} & \textbf{0.346} & \textbf{0.052} & \textbf{0.864} & \textbf{0.982} & \textbf{0.995} \\
    \midrule
    Auty~\etal~\cite{auty2023learning}  & 0.446 & 0.950 & 0.146 & 0.380 & 0.714 & 0.913 \\
    HoCLIP~\cite{zou2025high}           & \textbf{0.401} & 0.850 & 0.135 & 0.402 & 0.771 & 0.942 \\
    \rowcolor[HTML]{EFEFEF}
    \textbf{CLIP2Depth}                 & 0.431 & \textbf{0.535} & \textbf{0.096} & \textbf{0.627} & \textbf{0.908} & \textbf{0.968} \\
    \bottomrule
\end{tabular}
\label{tab:zeroshot}
\end{table}

\begin{figure}[t]
    \centering          
    \includegraphics[width=1.0\linewidth]{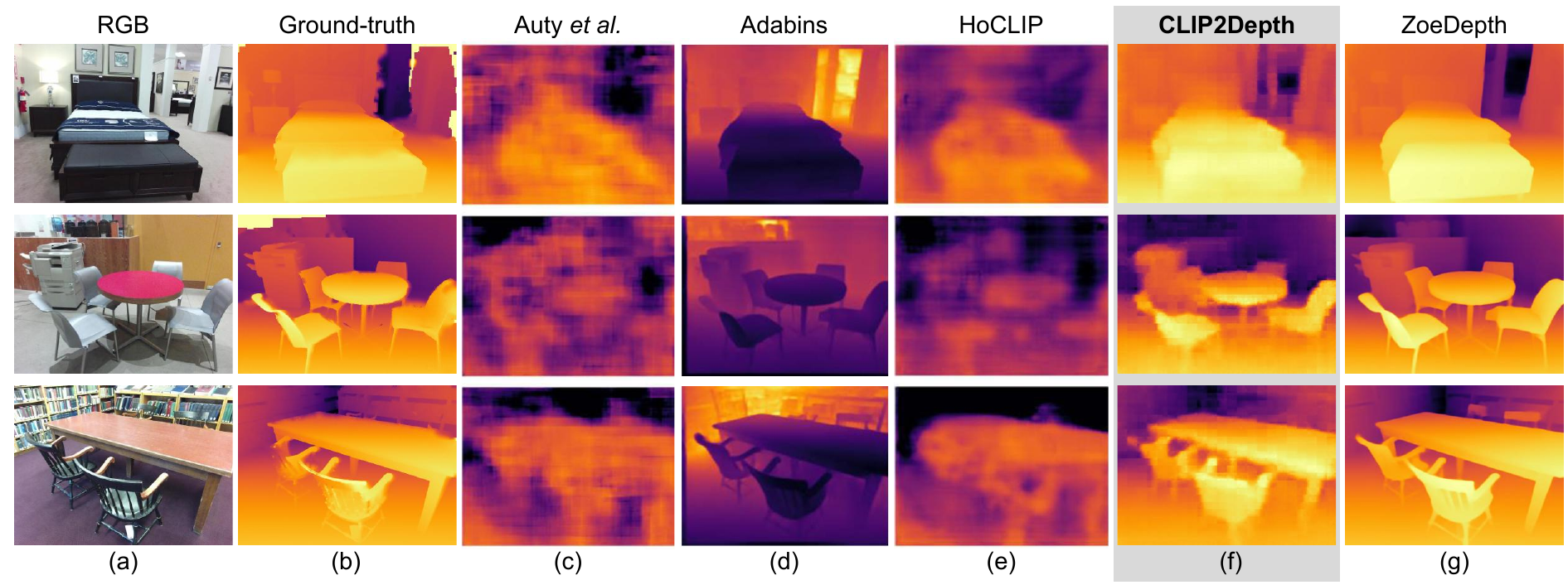} 
    \captionsetup{font=normalsize}
    \caption{Qualitative comparison of zero-shot generalization on the SUN RGB-D dataset: 
            (a) RGB input, (b) ground-truth, (c) Auty~\etal~\cite{auty2023learning}, (d) AdaBins~\cite{bhat2021adabins}, (e) HoCLIP~\cite{zou2025high}, (f) our CLIP2Depth, and (g) Zoedepth~\cite{bhat2023zoedepth}. 
            Columns (c–e) are taken from HoCLIP for fair comparison. 
            All depth maps are shown in absolute linear depth: (c, e) is presented with the \textit{inferno\_r} colormap, (d) with \textit{inferno}, and (b, f, g) are visualized with \textit{inferno\_r} using per-image min–max normalization.
            The comparison covers representative vision-only models (d, g) and CLIP-based approaches (c, e, f).}
    \label{fig:zeroshot}
\end{figure}

\subsection{Zero-shot Generalization}
\label{sec:zero-shot}
Beyond in-domain evaluation, it is also essential to examine how well the model generalizes to unseen datasets without additional training. 
In line with the observations in the previous section, where our method substantially outperforms prior CLIP-based approaches while remaining slightly behind large-scale vision-only models, we now evaluate its cross-dataset robustness. 
Specifically, the model is trained on the NYU Depth V2 dataset and tested on the SUN RGB-D dataset, both of which consist of challenging indoor scenes with distinct distributions. 
This setup allows us to simultaneously assess predictive accuracy and the ability to generalize across domains.

As summarized in \tabref{tab:zeroshot}, CLIP2Depth demonstrates clear improvements over prior CLIP-based methods in zero-shot evaluation. 
While its Abs Rel is not the lowest among all models, our method achieves the best performance across the remaining metrics (RMSE, $\log_{10}$, and $\delta^{1\text{--}3}$), highlighting its strong ability to preserve structural consistency and relative depth ordering under distribution shifts. 
Compared to large-scale vision-only baselines such as ZoeDepth~\cite{bhat2023zoedepth} and NeWCRFs~\cite{yuan2022neural}, CLIP2Depth still lags behind in overall absolute accuracy, yet delivers competitive quality with substantially fewer learnable parameters, underscoring the effectiveness of leveraging a frozen CLIP prior for generalizable depth estimation.

\figref{fig:zeroshot} further illustrates qualitative zero-shot results. 
For fairness, columns (c–e) are directly taken from the HoCLIP paper~\cite{zou2025high}.
CLIP2Depth produces sharper object boundaries and more coherent scene structures than prior CLIP-based approaches such as the method of Auty~\etal~\cite{auty2023learning} and HoCLIP~\cite{zou2025high}, while approaching the quality of vision-only baselines like ZoeDepth. 

\subsection{Consistency in Depth Estimation}
\label{sec:experiment_temporal}
In this section, we compare the depth consistencies of previous CLIP-based monocular depth approaches with ours.

\begin{figure}[t]
    \centering          
    \includegraphics[width=1.0\linewidth]{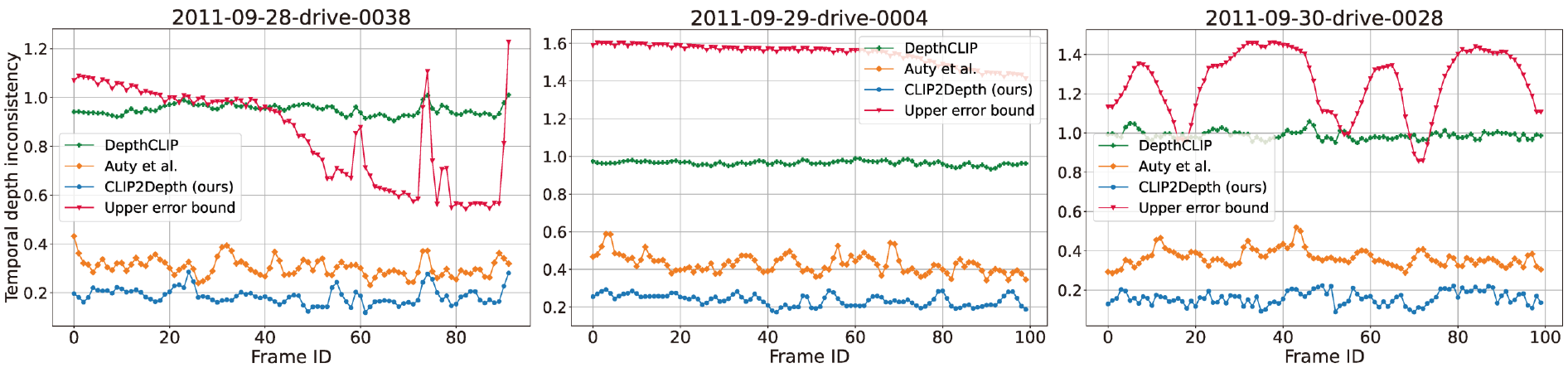} 
    \captionsetup{font=normalsize}
    \caption{Comparison of the temporal depth consistency on three selected KITTI~\cite{geiger2013vision} clips. 
    }
    \label{fig:temporal}
\end{figure}

\paragraph{\textbf{Temporal consistency}} 
We evaluate temporal consistency by selecting a video from the KITTI dataset~\cite{geiger2013vision}, as illustrated in \figref{fig:temporal}.
We compute~the temporal inconsistency of each pixel across all neighboring frames using the equation ${\vert \hat{d}_{t} - d_{t} \vert} / {\vert \hat{d}_{t} + d_{t} \vert}$,
where $\hat{d}_{t}$ refers to the re-projected depth map and $d_{t}$ denotes the ground-truth label of time $t$.
The relative camera motions for re-projection are obtained from the ground-truth~\cite{bian2019unsupervised}.
The upper error bound is set using predictions of a randomly initialized depth network proposed by Godard~\etal~\cite{godard2019digging}.
Each frame in four different video clips contains dynamic objects and camera motions. Our model demonstrates superior consistency with a much lower error rate compared to DepthCLIP~\cite{zhang2022can} and the method of Auty~\etal~\cite{auty2023learning}. 
DepthCLIP shows small variances but struggles with consistent depth values, often resulting in significant errors higher than the random upper bound. This indicates CLIP's limitations in continuous depth representation. Results from the method of Auty~\etal show better consistency, but their errors are still more than twice as high as ours. 
Despite the advantage from $32\times$ bilinear interpolation smoothing a large number of nearby depths to be similar, our approach outperforms both previous works with approximately 1M additional parameters, equivalent to a single linear layer. We conclude that CLIP2Depth effectively realigns CLIP's pre-trained knowledge for temporally consistent depth estimation.

\begin{figure}[t]
\small
    \centering
    \includegraphics[width=1.0\linewidth]{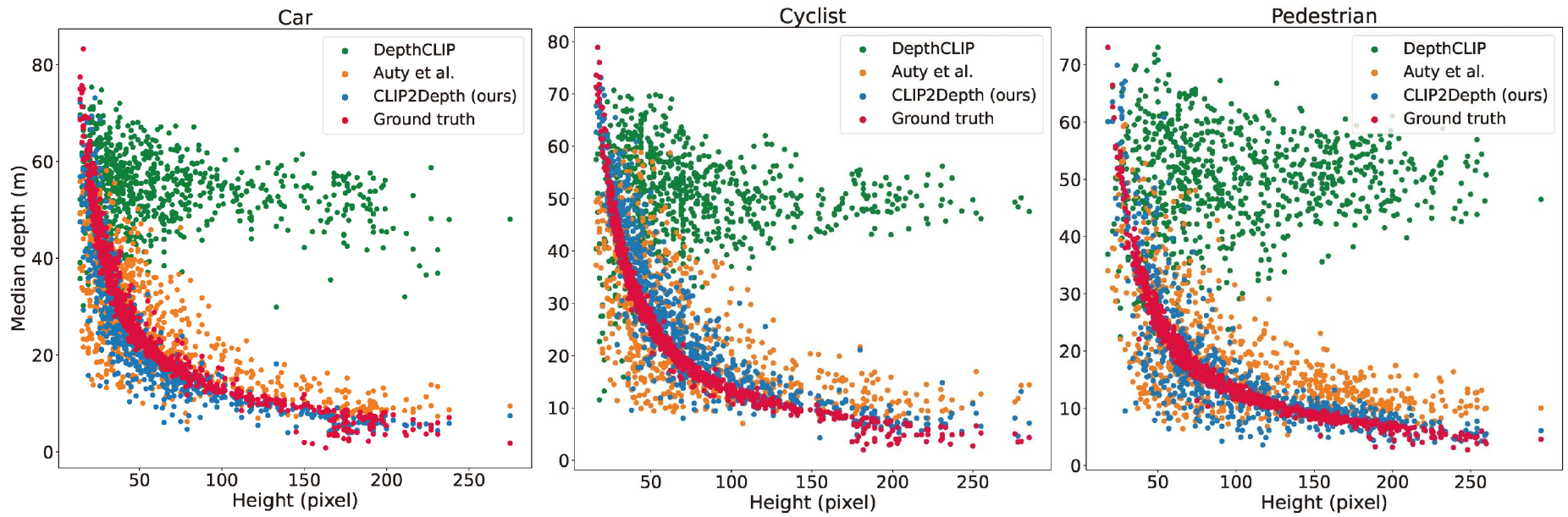}
    \captionsetup{font=normalsize}
    \caption{Comparison of spatial continuities between the predicted depths and object heights for the \texttt{Car}, \texttt{Cyclist}, and \texttt{Pedestrian} classes on the KITTI object detection dataset~\cite{Zhu_2019_ICCV}.
    }
    \label{fig:spatial}
\end{figure}

\paragraph{\textbf{Spatial continuity}}
\label{sec:experiment_continuous}
This experiment further demonstrates our superiority in preserving spatial continuity.
We verify that our model can consistently capture proper correlation between 2D scale (\emph{semantic cues}) and the associated 3D context (\emph{geometric cues}).
Using the three object categories from the KITTI object detection dataset~\cite{Zhu_2019_ICCV}: car, cyclist, and pedestrian,
we plot the~medians of the depths inside each bounding box and matching heights assigned to the labeled objects in $(x, y)$ pairs for comparison.

As shown in \figref{fig:spatial}, our model excels in preserving spatial continuity, aligning well with the distribution of ground-truth values.
In contrast, all other previous studies exhibit sporadic and discontinuous predictions.
DepthCLIP~\cite{zhang2022can} predicts depths within a limited range, so it struggles to infer the correct spatial correlation.
The method of Auty~\etal~\cite{auty2023learning} better estimates the distribution of ground-truth values; however, it still fails to capture the sharp semantic boundaries in its predicted depth maps, as shown in the qualitative results in \figref{fig:main}.
In comparison, our model consistently outperforms these methods in both spatial continuity and semantic sharpness, despite these comparative models having a considerable advantage in this experimental setting due to the depths smoothed by bilinear interpolation.

\subsection{Ablation Studies}
\label{sec:ablation}

\begin{figure*}[t!]
    \centering         
    \includegraphics[width=1.0\linewidth]{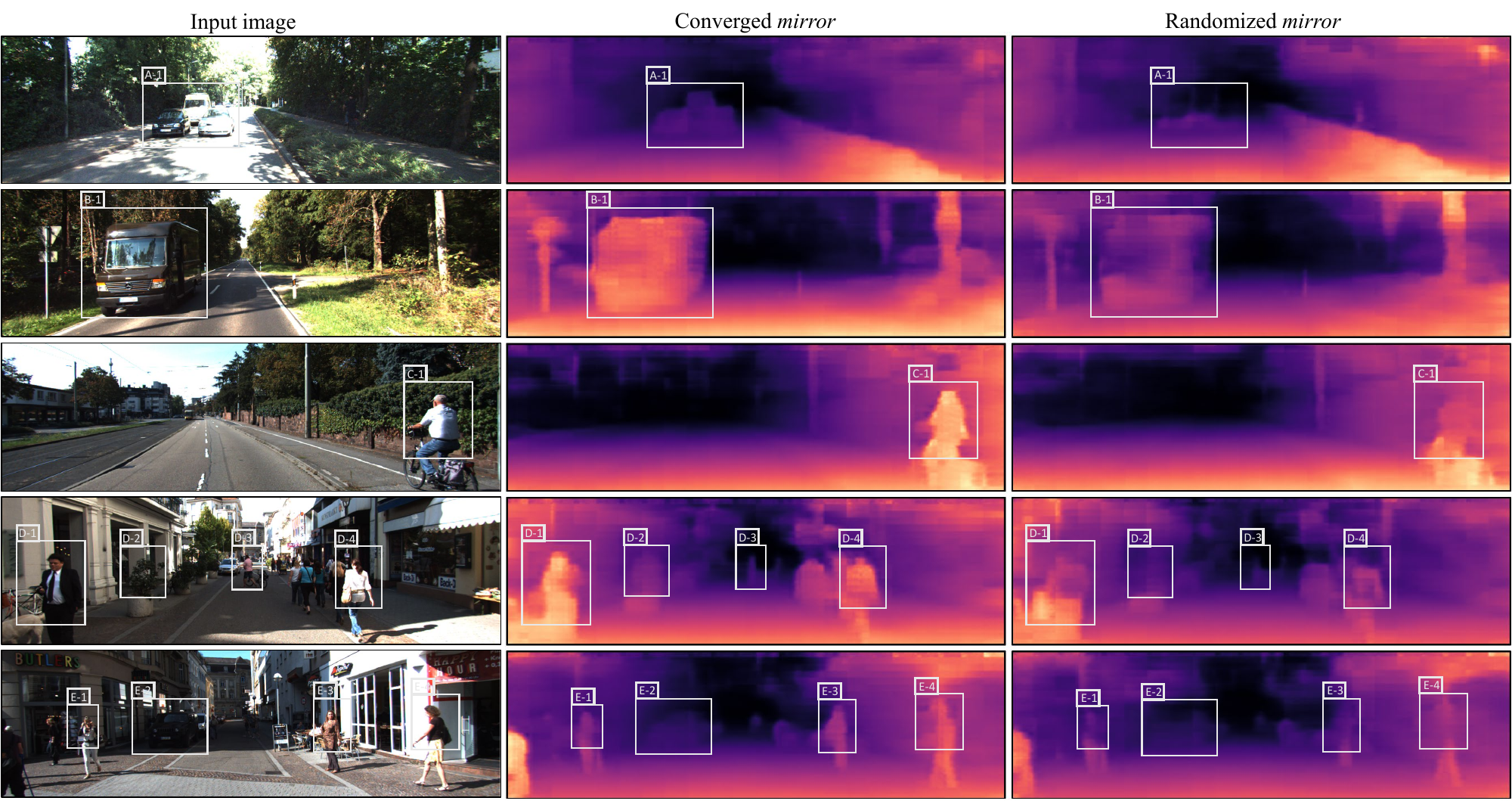} 
    \captionsetup{font=normalsize}
    \caption{Ablation study on the status of the \emph{mirror} in capturing objects for which semantic information serves a critical role. All results are absolute and inverse depth using the \emph{magma} colormap.}
    \label{fig:ablation_mirror}
\end{figure*}

For the ablation studies, we use the KITTI~\cite{geiger2013vision} dataset for all experiments (see \figref{fig:ablation_mirror} for reference). 
We analyze the effectiveness of each component in CLIP2Depth from three perspectives: 
(1) the effect of the \emph{mirror} in providing semantic guidance, 
(2) the effect of the image-text conditioning method, 
and (3) the effect of pre-training initialization. 
Each subsection provides a quantitative and qualitative comparison against ablated variants to highlight the strengths and necessity of our design choices.

\begin{table*}[t]
  \footnotesize
  \renewcommand*{\arraystretch}{1.8}
  \captionsetup{font=normalsize}
  \caption{
      Quantitative results on the KITTI~\cite{geiger2013vision} dataset comparing the characteristics of differently initialized \emph{mirror} embeddings.}
  \centering
  \begin{tabular}{lccccccc}
    \Xhline{\arrayrulewidth}
    \hline
    \textbf{Initialization} & $\bf Abs\,Rel\downarrow$ & $\bf Sq Rel\downarrow$ & $\bf RMSE\downarrow$ & $\mathrm{\bf{log_{10}}}\downarrow$ & $\bf \delta^{1}\uparrow$ & $\bf \delta^{2}\uparrow$ & $\bf \delta^{3}\uparrow$ \\
    \hline
    Randomized    & 0.0961   & 0.4628  & 3.6415 & 0.0420 & 0.8906 & 0.9770 & 0.9954 \\
     \cellcolor[HTML]{EFEFEF}Converged    
         & \cellcolor[HTML]{EFEFEF}\textbf{0.0739}
         & \cellcolor[HTML]{EFEFEF}\textbf{0.3028}  
         & \cellcolor[HTML]{EFEFEF}\textbf{2.9480}  
         & \cellcolor[HTML]{EFEFEF}\textbf{0.0322} 
         & \cellcolor[HTML]{EFEFEF}\textbf{0.9379}          
         & \cellcolor[HTML]{EFEFEF}\textbf{0.9899}          
         & \cellcolor[HTML]{EFEFEF}0.9979 \\
     Disrupted      & 0.0742 & 0.3070 & 2.9970 & 0.0325 & 0.9372 & 0.9898 & \textbf{0.9980}  \\
    \hline
    \Xhline{\arrayrulewidth}
  \end{tabular}
\label{tab:ablation_mirror}
\end{table*}

\subsubsection{Effect of Mirror}
In this subsection, we highlight the effect of \emph{mirror}, trained with the frozen CLIP text encoder, during depth estimation (see \tabref{tab:ablation_mirror}).
Despite being designed to distill the pre-trained semantic prior of the CLIP text encoder,
\emph{mirror} has few parameters ($\sim$30k) and is trained from scratch,
which could lead the entire CLIP2Depth framework to solely rely on the strong signals from the CLIP's image encoder and the decoder blocks with more parameters.
To clarify the role of the \emph{mirror} and verify that the model does not disregard its supervision during training, we conduct three ablation experiments.
As shown in \tabref{tab:ablation_mirror}, we compare our model (second row, labeled ``Converged'') with a variant in which the \emph{mirror} is randomized after training (first row, labeled ``Randomized'').  
The performance gap between the two indicates that the learned \emph{mirror} contributes significantly to depth prediction quality.
We further validate this finding by introducing a third variant (``Disrupted''), in which the \emph{mirror} are re-randomized at every training step.  
This forces the model to learn without relying on any consistent semantic guidance from the \emph{mirror}.  
Although the overall performance of the ``Converged'' and ``Disrupted'' models is similar, the ``Converged'' model suffers a notable degradation when its trained \emph{mirror} is subsequently randomized, confirming that it had indeed learned to depend on the \emph{mirror}'s semantic guidance.

This trend consistently appears in the ablation result in \figref{fig:ablation_mirror}:
the second column shows the qualitative results of the model (``Converged"), and the third column shows the randomized ablation.
The white-colored bounding boxes label the objects in the scene, where understanding semantic context is crucial for their detection. 
The ablated model with randomized \emph{mirror} fails to predict the depth boundary of the annotated objects, indicating that \emph{mirror} effectively identifies semantic objects in given scenes.
Notably, although \emph{mirror} is used for depth estimation across the entire image,
its main contribution lies in capturing semantic cues, aligning with its original design objective.

\begin{table*}[t]
  \renewcommand*{\arraystretch}{1.8}
  \footnotesize
  \captionsetup{font=normalsize}
  \caption{
      Ablation results on the conditioning method for depth estimation used in the method of Auty~\etal~\cite{auty2023learning} and ours.}
  \centering
  \begin{tabular}{lccccccc}
    \Xhline{\arrayrulewidth}
    \hline
    \textbf{Conditioning} & $\bf Abs\,Rel\downarrow$ & $\bf Sq Rel\downarrow$ & $\bf RMSE\downarrow$ & $\mathrm{\bf{log_{10}}}\downarrow$ & $\bf \delta^{1}\uparrow$ & $\bf \delta^{2}\uparrow$ & $\bf \delta^{3}\uparrow$ \\
    \hline
    Auty~\etal~\cite{auty2023learning}    & 0.0802 & 0.3502 & 3.1390 & 0.0348 & 0.9239 & 0.9873 & 0.9974 \\
     \cellcolor[HTML]{EFEFEF} \textbf{CLIP2Depth} 
         & \cellcolor[HTML]{EFEFEF}\textbf{0.0768} 
         & \cellcolor[HTML]{EFEFEF}\textbf{0.3291}  
         & \cellcolor[HTML]{EFEFEF}\textbf{3.1267}  
         & \cellcolor[HTML]{EFEFEF}\textbf{0.0338} 
         & \cellcolor[HTML]{EFEFEF}\textbf{0.9287}          
         & \cellcolor[HTML]{EFEFEF}\textbf{0.9883}           
         & \cellcolor[HTML]{EFEFEF}\textbf{0.9977} \\
    \hline
    \Xhline{\arrayrulewidth}
  \end{tabular}
    \label{tab:ablation_modulation}
\end{table*}

\subsubsection{Effect of Conditioning Method} Unlike in previous CLIP-based monocular depth estimation methods that correlate images with distance-related prompts by computing pairwise similarity scores~\cite{zhang2022can,auty2023learning,hu2023learning}, our framework modulates images using a single, non-human language prompt as the input of a FiLM~\cite{perez2018film} block. This design choice, utilizing \emph{mirror}, is more effective for correcting misaligned correlation in the depth domain.

To accurately assess the effects of image conditioning methods, we jointly trained a model reproduced from Auty~\etal~\cite{auty2023learning}, which uses the shallow transformer and deconvolution blocks of the CLIPSeg decoder~\cite{Luddecke_2022_CVPR} without the FiLM block for comparison with our method.
For fairness, both models used 64 depth tokens and mirror embeddings and were trained from scratch without pre-training on any CLIPSeg component. The quantitative results in \tabref{tab:ablation_modulation}
show a clear performance advantage in every metric, strongly supporting our intuition that querying depth cues by reflecting the entire image in a single non-human language prompt is more effective than combining similarities with multiple prompts into a single depth value.

\subsubsection{Effect of pre-training} 
We compare the effects of different pre-training approaches. See Table~\ref{tab:init_ablation} for quantitative results: The model in the first row is trained from scratch, without pre-training (``Randomized''), and the second row is fine-tuned from weights pre-trained on the PhraseCut~\cite{wu2020phrasecut} dataset (CLIPSeg~\cite{Luddecke_2022_CVPR}) as introduced in our CLIP2Depth framework. 
As briefly stated in Section~\ref{sec:method}, we design our framework to leverage the prior knowledge on semantic segmentation of pre-trained CLIPSeg. 
Our fine-tuned model outperforms the randomized initialization model on all metrics, validating the effectiveness of transferring semantic segmentation knowledge.
Based on this, we suggest that transferring knowledge from semantic segmentation can be particularly effective in adjusting incorrect image-text correlation biases in vision-language foundation models. This is achieved by first training a learnable component for semantic segmentation with a frozen vision-language model, and then fine-tuning it for target domains.

\begin{table*}[t]
  \renewcommand*{\arraystretch}{1.5}
  \footnotesize
  \captionsetup{font=normalsize}
  \caption{Ablation results comparing initialization strategies (Randomized vs. CLIPSeg) for depth estimation.}
  \centering
  \begin{tabular}{lccccccc}
    \Xhline{\arrayrulewidth}
    \hline
    \textbf{Initialization} & $\bf Abs\,Rel\downarrow$ & $\bf Sq Rel\downarrow$ & $\bf RMSE\downarrow$ & $\mathrm{\bf{log_{10}}}\downarrow$ & $\bf \delta^{1}\uparrow$ & $\bf \delta^{2}\uparrow$ & $\bf \delta^{3}\uparrow$ \\
    \hline
    Randomized
         & 0.077 
         & 0.329 
         & 3.127 
         & 0.034 
         & 0.929 
         & 0.988 
         & \textbf{0.998} \\
         
    \cellcolor[HTML]{EFEFEF} CLIPSeg~\cite{Luddecke_2022_CVPR}
         & \cellcolor[HTML]{EFEFEF}\textbf{0.074} 
         & \cellcolor[HTML]{EFEFEF}\textbf{0.303}  
         & \cellcolor[HTML]{EFEFEF}\textbf{2.948}  
         & \cellcolor[HTML]{EFEFEF}\textbf{0.032} 
         & \cellcolor[HTML]{EFEFEF}\textbf{0.938}          
         & \cellcolor[HTML]{EFEFEF}\textbf{0.990}           
         & \cellcolor[HTML]{EFEFEF}\textbf{0.998} \\
    \hline
    \Xhline{\arrayrulewidth}
  \end{tabular}
  \label{tab:init_ablation}
\end{table*}

\section{Limitations}
While the proposed CLIP2Depth framework demonstrates significant improvements in monocular depth estimation, it does not yet reach the state-of-the-art performance achieved by CLIP in other applications, such as semantic segmentation and generative modeling. 
This suggests that inherent limitations remain in fully leveraging the pretrained vision-language prior of CLIP for depth estimation, particularly due to the distinct nature of geometric reasoning required for this task.

Moreover, although the introduction of the non-human language prompt embedding, \emph{mirror}, provides an effective mechanism for adapting CLIP to depth estimation, its effectiveness may heavily depend on the presence of semantically rich content within the scene. This raises potential concerns about generalization in environments where geometric structures dominate and semantic cues are sparse.

Also, replacing natural language prompts with non-human language embeddings could compromise CLIP's inherent open-vocabulary reasoning capabilities, potentially limiting its broader applicability to general vision-language tasks.

\section{Future Works}
Future work should focus on enhancing the generalizability and robustness of the proposed method, potentially by integrating advanced fine-tuning techniques or hybrid approaches to narrow the performance gap with leading vision models. 
Additionally, exploring strategies to effectively incorporate open-vocabulary capabilities while maintaining the advantages of non-human language prompting could further expand the applicability and performance of CLIP2Depth across a broader range of tasks.
Furthermore, investigating methods to improve cross-domain generalization, particularly in scenes with limited semantic cues or diverse geometric structures, would further validate the broader applicability of our approach.

\section{Conclusion}
In this work, we demonstrate that CLIP can be effectively adapted for monocular depth estimation, despite the suboptimal nature of its pretrained prior, and without requiring fine-tuning or manual engineering of its subword embeddings. 
Our proposed CLIP2Depth framework distills the semantic priors of the frozen CLIP text encoder into a compact non-human language prompt, achieving depth estimation performance that is comparable to previous state-of-the-art methods. 
Specifically, CLIP2Depth reduces the absolute relative error by 68.7\% on NYU Depth v2 and 75.6\% on KITTI compared to the method of Auty~\etal, the representative CLIP-based baseline.

In addition, ablation studies on the \emph{mirror} embedding highlight its clear role in capturing object depths, particularly in regions where semantic cues are essential for accurate prediction. Beyond improvements in accuracy, our method addresses CLIP’s inherent limitations in spatial reasoning by introducing a lightweight, task-specific prompting mechanism. This approach enhances both spatial and temporal consistency while maintaining low model complexity and avoiding any modification of the pretrained CLIP backbone.
We hope that this work provides a new perspective on adapting vision-language models to geometric tasks, encouraging further exploration in this direction.

\section{Acknowledgements}
This research was financially supported by the Ministry of Trade, Industry and Energy, Korea, under the “Regional Innovation Cluster Development Program(R\&D, p0025331)” supervised by the Korea Institute for Advancement of Technology(KIAT) (50\%).
This work was supported by the National Research Foundation of Korea(NRF) grant funded by the Korea government(MSIT)(RS-2024-00457860).   
This research was supported by the Korea Astronomy and Space Science Institute under the R\&D program(Project No. 2024-1-904-03) supervised by the Korea AeroSpace Administration.

\bibliographystyle{elsarticle-num} 
\bibliography{custom}

\end{document}